\documentclass[10pt, a4paper]{article}
\usepackage{graphicx}
\usepackage[final]{lrec-coling2024} 

\title{Training a Bilingual Language Model by Mapping \\Tokens onto a Shared Character Space}

\name{Aviad Rom, Kfir Bar} 

\address{The Data Science Institute, Reichman University, Herzliya, Israel\\
\{aviad.rom,kfir.bar\}@post.idc.ac.il\\}

\abstract{
We train a bilingual Arabic-Hebrew language model using a transliterated version of Arabic texts in Hebrew, to ensure both languages are represented in the same script. Given the morphological, structural similarities, and the extensive number of cognates shared among Arabic and Hebrew, we assess the performance of a language model that employs a unified script for both languages, on machine translation which requires cross-lingual knowledge. The results are promising: our model outperforms a contrasting model which keeps the Arabic texts in the Arabic script, demonstrating the efficacy of the transliteration step. 
    Despite being trained on a dataset approximately 60\% smaller than that of other existing language models, our model appears to deliver comparable performance in machine translation across both translation directions.
 \\ \newline  }

\begin{document}

\maketitleabstract
\section{Introduction}
Pre-trained language models have become essential for state-of-the-art performance in mono- and multilingual natural language processing (NLP) tasks.
They are pre-trained once and can then be fine-tuned for various downstream NLP tasks.
It has been shown that language models generalize better on multilingual tasks when the target languages share structural similarity, possibly due to script similarity \citep{DBLP:conf/iclr/KWMR20}.
Typically, language models are trained on sequences of tokens that often correspond to words and subword components. 

Arabic and Hebrew are two Semitic languages that share similar morphological structures in the composition of their words, but use distinct scripts for their written forms.
The Hebrew script primarily serves Hebrew, but is also employed in various other languages used by the Jewish population. 
These languages include Yiddish (or ``Judeo-German''), Ladino (or ``Judeo-Spanish''), and Judeo-Arabic, which comprises a cluster of Arabic dialects spoken and written by Jewish communities residing in Arab nations.
To some extent, Judeo-Arabic can be perceived as an Arabic variant written in the Hebrew script.
Most of the vocabulary in Judeo-Arabic consists of Arabic words that have been transliterated into the Hebrew script.

Words in two languages that share similar meanings, spellings, and pronunciations are known as cognates.
Arabic and Hebrew cognates share similar meanings and spellings despite different scripts.
The pronunciation of such cognates are not necessarily the same.
Numerous lexicons have been created to record these cognates. 
One of those lexicons\footnote{\url{https://seveleu.com/pages/semitic-syntax-morpho/comparative-sem}} lists a total of 915 Hebrew-Arabic spelling equivalents, of which 435 have been identified as authentic cognates, signifying that they possess identical meanings.
Analyzing a parallel Hebrew-Arabic corpus, named Kol Zchut\footnote{\url{https://releases.iahlt.org/}} using this lexicon, we found instances of those cognates in about 50\% of the sentences.

The purpose of this work is to take advantage of the potentially high frequency of cognates in Arabic and Hebrew in building a bilingual language model using only one script. 
Subsequently, the model will be fine-tuned on NLP tasks, such as machine translation, which can benefit from the innate bilingual proficiency to achieve better results.
To ensure that cognates are mapped onto a consistent character space, the model uses Arabic texts that are transliterated into the Hebrew script, which mimics the writing system used in Judeo-Arabic.
We call this new language model \texttt{HeArBERT}.\footnote{\url{https://huggingface.co/aviadrom/HeArBERT}}

We test our new model on machine translation, which is considered a downstream task requiring knowledge in two languages, and report on some promising results. 
In summary, the primary contributions of our work are:
(1) we release a new bilingual Arabic-Hebrew language model; and, (2) we show that pre-training a bilingual language model on transliterated texts, as a way for aligning tokens onto the same character space, is beneficial for machine translation.

\section{Related Work}


\citet{DBLP:conf/iclr/KWMR20} have suggested that structural similarity of languages is essential for language model's multilingual generalization capabilities.
Their suggestion was further discussed by \citet{dufter-schutze-2020-identifying}, who highlighted the essential components for a model to possess ``multilinguality'', and show that the order of the words in the sentence is key to the model's cross-lingual generalization capabilities.
mBERT, as introduced by \citet{devlin-etal-2019-bert}, was a pioneering language model that encompassed multiple languages, including Arabic and Hebrew.
However, both Arabic and Hebrew are significantly under-represented in the pre-training data, resulting in inferior performance compared to the equivalent monolingual models on various downstream tasks \citep{antoun-etal-2020-arabert,lan-etal-2020-empirical,chriqui2021hebert,seker-etal-2022-alephbert}. 
GigaBERT \cite{lan-etal-2020-empirical} is another multilingual model, trained for English and Arabic.
However, the best results for most of the known NLP tasks are typically achieved by one of the large monolingual models in both Arabic and Hebrew.
CAMeLBERT \citep{inoue-etal-2021-interplay}, is one of those models. 
It is trained on texts written in Modern Standard Arabic (MSA), Classical Arabic, as well as dialectal variants.
In the realm of Hebrew language models, AlephBERT \citep{seker-etal-2022-alephbert} stands out as one of the leading performers, alongside others like HeBERT \citep{chriqui2021hebert}.
Among other datasets, the monolingual models mentioned above use the relevant parts of the OSCAR dataset \citep{ortiz-suarez-etal-2020-monolingual} for training. 
Our model relies solely on the OSCAR dataset for both Hebrew and Arabic, resulting in a considerably smaller total number of words for each language in comparison to the existing monolingual language models.

The effect of transliteration on cross-lingual generalization were discussed previously in \citep{dhamecha-etal-2021-role,chau-smith-2021-specializing} and more recently in \citep{moosa-etal-2023-transliteration,purkayastha2023romanization}. 
None of these works study the languages of our focus: Arabic and Hebrew.
\citet{dhamecha-etal-2021-role} focused on languages from the Indo-Aryan family, which has been studied before for cross-lingual generalization and also has several publicly available multilingual models. 
To the best of our knowledge, our work is first to study generalization between Arabic and Hebrew and no multilingual masked language models that include both languages have been published apart from mBERT. 

\citet{chau-smith-2021-specializing} address the generalization from high- to low-resourced languages. 
However, both Arabic and Hebrew are currently considered medium- to high-resourced languages. 
Furthermore, their evaluation focuses solely on token-level classification tasks, such as dependency parsing and part-of-speech tagging, whereas our evaluation targets machine translation, a sequence-to-sequence bilingual task.

\citet{purkayastha2023romanization} employ Romanization for transliteration, whereas we transliterate Arabic into the Hebrew script. Analogous to \citet{chau-smith-2021-specializing}, their evaluation centers on token-level classification tasks, which are not addressed in our work.

\section{Methodology}

\label{arabic-hebrew-transformation}
We begin by pre-training a new language model using texts written in both Arabic and Hebrew. 
This model, named \texttt{HeArBERT}, is subsequently fine-tuned to enhance performance in machine translation between Arabic and Hebrew.

For pre-training, we utilize the de-duplicated Arabic ($\sim$3B words) and Hebrew ($\sim$1B words) versions of the OSCAR dataset \citep{ortiz-suarez-etal-2020-monolingual}.
In this work, we aim to measure the impact of normalizing all texts to a shared script, so that cognates can be unified under the same token representation.
Therefore, we transliterate the Arabic texts into the Hebrew script as a preprocessing step for both training and testing.
Our transliteration procedure is designed following most of the guidelines published by \textit{The Academy of the Hebrew Language} who has defined a Hebrew mapping for every Arabic letter\footnote{\url{https://hebrew-academy.org.il/wp-content/uploads/taatik-aravit-ivrit-1.pdf}}, and the mapping provided in \citep{Terner}. 
Only Arabic letters are converted to their Hebrew equivalents, while non-Arabic characters remain unchanged.
Our implementation is based on a simple lookup table, executed letter by letter, which is composed of the two mappings mentioned above, as shown in Appendix \ref{sec-transliteration-map-appendix}.

For evaluation, we independently train the model twice: once with the transliteration step and once without. 
We subsequently compare the performance of these two versions when fine-tuned on a downstream machine translation test set.

Our model is based on the original BERT-base architecture.
We train a WordPiece tokenizer with a vocabulary size of $30,000$, limiting its accepted alphabet size to $100$. 
This approach encourages the learning of tokens common to both languages, allowing the tokenizer to focus on content rather than on special characters not inherent to either language.
We choose to train only for the masked language model (MLM) task employed originally in BERT, ignoring the next-sentence-prediction component, as it has previously been proven less effective \citep{liu2019roberta}.
Overall, we trained each model for the duration of 10 epochs, over the course of approximately 3 weeks, using 4 Nvidia RTX 3090 GPUs.

Fine-tuning HeArBERT is done similar to fine-tuning the original BERT model, except for the addition of the transliteration step of Arabic letters that takes place prior to tokenization.
In this pre-processing step, all non-Arabic letters remain intact, while Arabic letters are transliterated into their Hebrew equivalents, as described above.

\begin{figure}[!ht]
\begin{center}
\includegraphics[scale=0.35]{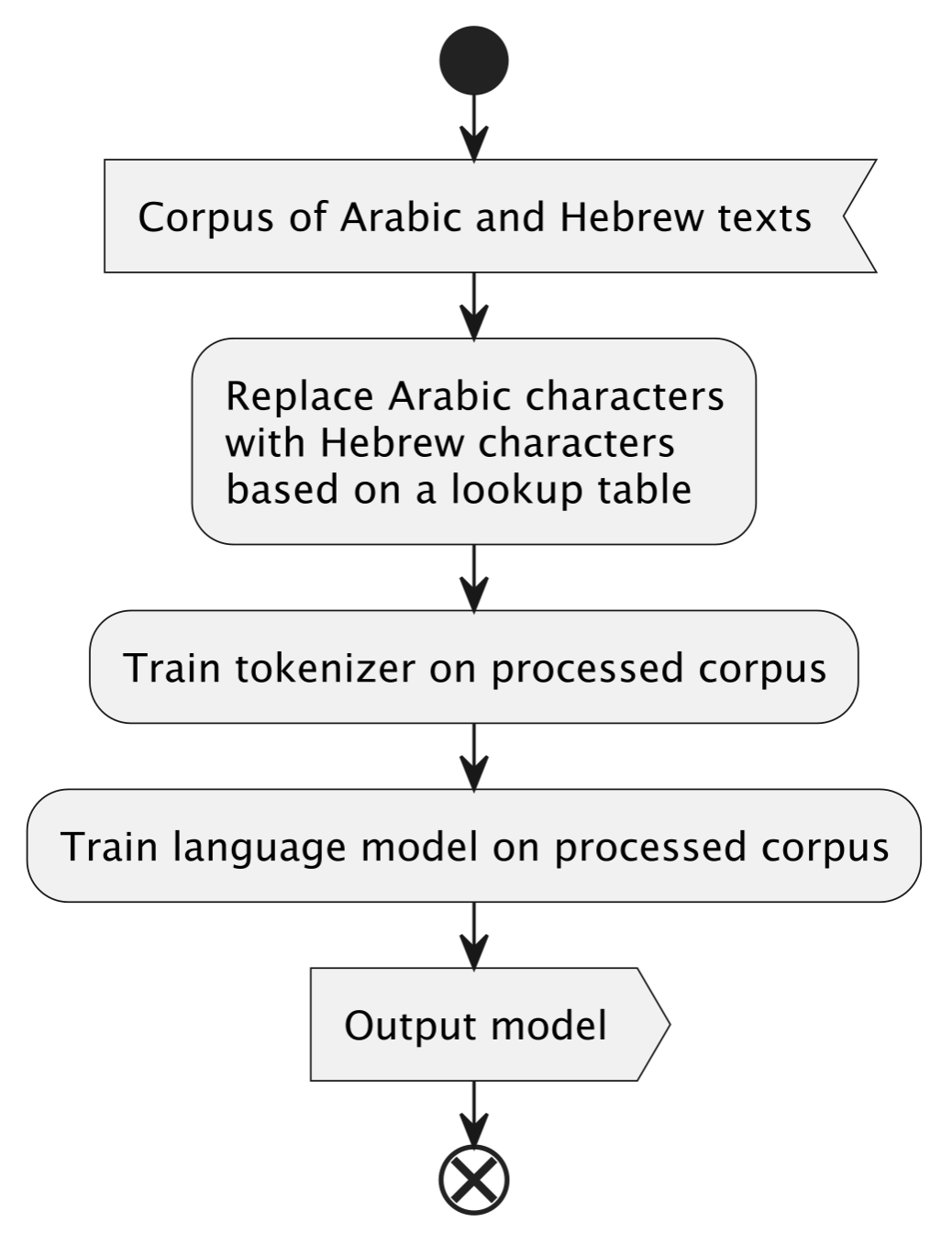} 
\caption{Pre-training process of HeArBERT.}
\label{pretrain_flow}
\end{center}
\end{figure}

The preprocssing and pre-training process of HeArBERT is depicted in Figure \ref{pretrain_flow}
\section{Experimental Settings}

\paragraph{\textbf{Machine Translation.}}Our machine-translation architecture is based on a simple encoder-decoder framework, which we initialize using weights of the BERT model in focus.\footnote{We use HuggingFace's \texttt{EncoderDecoderModel}.}
For example, if we want our model to translate from Hebrew to Arabic, we might initialize the encoder with the model weights of mBERT and the decoder with the weights of CAMelBERT.
To fine-tune the model on machine translation, we use the new ``Kol Zchut'' (in English, ``All Rights'') Hebrew-Arabic parallel corpus\footnote{\url{https://releases.iahlt.org/}} which contains over 4,000 parallel articles in the civil-legal domain, corresponding to 140,000 sentence-pairs in Arabic and Hebrew containing 2.13M and 1.8M words respectively.
To the best of our knowledge, ours is the first work to report machine translation results using this resource; therefore, no established baseline or benchmark exists for comparison.
As the corpus is provided without an official train/test split, we apply a random split with $80\%$ of the data being allocated for training and the remaining $20\%$ for testing, using the \texttt{train\_test\_split} function of \texttt{scikit-Learn} with a random seed of 42.
We evaluate our HeArBERT-based translation against an equivalent system initialized using other models. 
The standard BLEU metric \citep{papineni-etal-2002-bleu} is employed to contrast the system's generated translation with the sole reference translation present in the corpus. 
Each system is fine-tuned for the duration of ten epochs, and we report the best performance observed across all epochs.

\paragraph{\textbf{Baseline Language Models.}}

To contrast HeArBERT with an equivalent model trained on texts in both Arabic and Hebrew scripts, we pre-train another model and tokenizer following the same procedure, but without the transliteration preprocessing for the Arabic data. 
This model is denoted with a subscript "NT" (no transliteration): \texttt{HeArBERT$_{NT}$}.

We compare our model with a number of existing models.
The initial model, mBERT, was originally pre-trained on a range of languages, including both Arabic and Hebrew. 
We also chose a couple of monolingual Arabic language models, with specific versions from Hugging Face provided in a footnote.
Specifically, we use CAMeLBERT\footnote{\texttt{CAMeL-Lab/bert-base-arabic-camelbert-mix}} and GigaBERT\footnote{\texttt{lanwuwei/GigaBERT-v4-Arabic-and-English}}.
In some experiments, we adopt a technique inspired by \citet{rom-bar-2021-supporting}. 
This involves expanding the vocabulary of an existing Arabic language-model's tokenizer by appending a Hebrew-transliterated version of each Arabic token and associating it with the original token identifier.
We denote such extended models by adding ``ET'' (extended tokenizer) to the model name.

All these models share the same architecture size as our proposed model.

\begin{table*}[hbt!]
    \centering
    \begin{tabular}{l|l|r||l|l|r}
        \hline
        \multicolumn{3}{c}{\textbf{Arabic-to-Hebrew}} & \multicolumn{3}{c}{\textbf{Hebrew-to-Arabic}} \\
        \hline
        \textbf{Encoder} & \textbf{Decoder} &  \textbf{BLEU} & \textbf{Encoder} & \textbf{Decoder} &  \textbf{BLEU}  \\
        \hline
        mBERT & mBERT & 15.59 & mBERT & mBERT & 11.48 \\
        CAMeLBERT & CAMeLBERT$_{ET}$ & 12.47 & CAMeLBERT$_{ET}$ & CAMeLBERT & 16.86 \\
        CAMeLBERT$_{ET}$ & CAMeLBERT$_{ET}$ & 12.66 & CAMeLBERT$_{ET}$ & CAMeLBERT$_{ET}$ & 17.15 \\
        \hline
        HeArBERT (ours) & HeArBERT (ours) & 24.97 & HeArBERT & HeArBERT & 18.92\\
        GigaBERT & HeArBERT & \textbf{25.28} & HeArBERT & GigaBERT & \textbf{21.17}\\
        CAMeLBERT & HeArBERT & 23.69 & HeArBERT & CAMeLBERT & 19.41 \\
        \hline
        HeArBERT$_{NT}$ & HeArBERT$_{NT}$ & 23.97 & HeArBERT$_{NT}$ & HeArBERT$_{NT}$ & 18.32\\
        GigaBERT & HeArBERT$_{NT}$ & 23.73 & HeArBERT$_{NT}$ & GigaBERT & 21.00\\
        CAMeLBERT & HeArBERT$_{NT}$ & 22.76 & HeArBERT$_{NT}$ & CAMeLBERT & 19.05 \\
        \hline
    \end{tabular}
    \caption{Machine translation performance (BLEU scores on the Kol Zchut test set).}
    \label{tab:translation}
\end{table*}

\section{Results}
\label{sec:results}
The results are summarized in Table \ref{tab:translation}.
The table rows are organized into three distinct groups. 
The first group features combinations of various baseline models. 
The second group showcases combinations incorporating our proprietary HeArBERT model, while the third group highlights combinations involving the contrasting HeArBERT$_{NT}$ model.
We train multiple machine-translation combinations, in both directions Arabic-to-Hebrew and Hebrew-to-Arabic, based on the same encoder-decoder architecture, initialized with different language model combinations.
We assign various combinations of language models to the encoder and decoder components, ensuring that the selected models align with the respective source and target languages.
In other words, we make sure that the language model which we use to initialize the encoder, is compatible with the system's source language.

Since mBERT and CAMeLBERT$_{ET}$, as well as our two models HeArBERT and HeArBERT$_{NT}$, can potentially handle both languages, we experiment with combinations where each of them is assigned to both, the encoder and decoder components at the same time.

The results demonstrate that the second group, which utilizes our HeArBERT to initialize a Hebrew decoder, surpasses the performance of all other combinations. 
Notably, the pairing of GigaBERT with HeArBERT is the standout performer across both directions.
It surpasses the performance of using HeArBERT for initializing the Arabic encoder by only a few minor points.
The performance of HeArBERT$_{NT}$, as reported in the third group is consistently lower than HeArBERT in both directions. 
The difference seems to be more significant in the Arabic-to-Hebrew direction.
Overall, the results show that transliterating the Arabic training texts into the Hebrew script as a pre-processing step is beneficial for an Arabic-to-Hebrew machine translation system.
The impact of the transliteration proves to be less pronounced in the reverse translation direction.

Using the extended (ET) version of CAMeLBERT has a reasonable performance.
However, it performs much worse than the best model in both directions, indicating that extending the vocabulary with transliterated Arabic tokens does contribute to better capturing the meaning of Hebrew tokens in context.
This implies that joint pre-training on both languages is essential for achieving a more robust language representation.

\section{Conclusion}
Arabic and Hebrew, both Semitic languages, display inherent structural resemblances and possess shared cognates. 
In an endeavor to allow a bilingual language model to recognize these cognates, we introduced a novel language model tailored for Arabic and Hebrew, wherein the Arabic text is transliterated into the Hebrew script prior to both pre-training and fine-tuning.
We contrast our model by training another language model on the identical dataset but without the transliteration preprocessing step, in order to assess the impact of transliteration.
We fine-tuned our model for the machine translation task, yielding promising outcomes. These results suggest that the transliteration step offers tangible benefits to the translation process.

Comparing to the translation combination involving other language models, we see comparable results; this is encouraging given that the training data we utilized for pre-training the model is approximately 60\% smaller than theirs.

As a future avenue of research, we intend to train the model on an expanded dataset and explore scaling its architecture. 
In this study, our emphasis was on a transformer encoder. 
We are keen to investigate the effects of implementing transliteration within a decoder architecture, once such a model becomes available for Hebrew.

\section*{Limitations}
The transliteration algorithm from Arabic to Hebrew is based a simple deterministic lookup table. 
However, sometimes the transliteration is not that straight forward, and this simple algorithm generates some odd rendering, which we would like to fix.
For example, our algorithm does not place a final-form letter at the end of the Arabic word in Hebrew.
Another challenge with transliteration into Hebrew is that for some words a Hebrew writer may choose to omit long vowel characters and the readers will still be able to understand the word.
This phenomenon is referred to as ``Ktiv Hasser'' in Hebrew. 
Yet, there exists a preference for certain word representations over others.
This inconsistency makes it more challenging for aligning the transliterated Arabic words to their cognates in the way they are written.
Our transliteration algorithm always renders the Arabic word following the Arabic letters, which may be different than how this word is typically written in Hebrew.

Another limitation is the relatively small size of the dataset which we used for pre-training the language model, comparing to other existing language models of the same architecture size.




\section{Bibliographical References}\label{sec:reference}

\bibliographystyle{lrec-coling2024-natbib}
\bibliography{lrec-coling2024-example}
\appendix

\section{Transliteration Table}
\label{sec-transliteration-map-appendix}
In Table \ref{tab:transliteration-mapping} we provide the transliteration table that we use for transliterating Arabic texts into the Hebrew script as a pre-processing step in HeArBERT and in the tokenizer extension for CAMeLBERT$_{ET}$.

\begin{table*}[htp!]
    \centering
    \includegraphics[width=400pt]{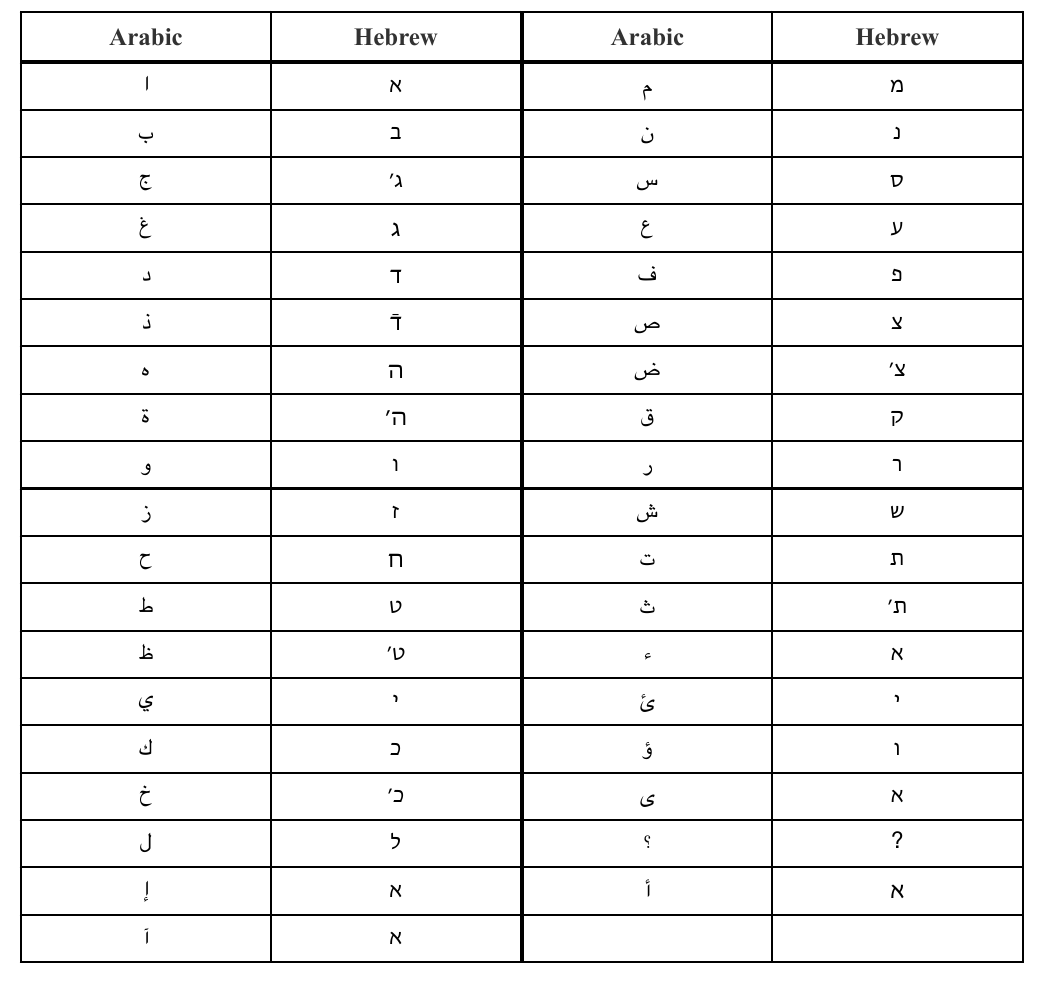}
    \caption{Character mapping used for Arabic-to-Hebrew transliteration.}
    \label{tab:transliteration-mapping}
\end{table*}

\end{document}